\documentclass[runningheads]{llncs}

\usepackage{eccv}

\usepackage{eccvabbrv}
\usepackage{multirow} 
\usepackage{amsmath}
\usepackage{colortbl}
\usepackage{pifont}
\newcommand{\cmark}{\ding{51}}% 对号
\newcommand{\xmark}{\ding{55}}% 错号
\usepackage{subcaption}
\usepackage{colortbl}
\usepackage{algorithm}
\usepackage{algpseudocode}
\usepackage{multirow}
\usepackage{wrapfig,lipsum,booktabs}
\usepackage[misc]{ifsym}
\usepackage[accsupp]{axessibility}  % Improves PDF readability for those with disabilities.
\usepackage{lipsum}
\newcommand\blfootnote[1]{%
  \begingroup
  \renewcommand\thefootnote{}\footnote{#1}%
  \addtocounter{footnote}{-1}%
  \endgroup
}

\usepackage{algorithm}
\usepackage{etoolbox}
\makeatletter
\AfterEndEnvironment{algorithm}{\let\@algcomment\relax}
\AtEndEnvironment{algorithm}{\kern2pt\hrule\relax\vskip3pt\@algcomment}
\let\@algcomment\relax
\newcommand\algcomment[1]{\def\@algcomment{\footnotesize#1}}
\renewcommand\fs@ruled{\def\@fs@cfont{\bfseries}\let\@fs@capt\floatc@ruled
  \def\@fs@pre{\hrule height.8pt depth0pt \kern2pt}%
  \def\@fs@post{}%
  \def\@fs@mid{\kern2pt\hrule\kern2pt}%
  \let\@fs@iftopcapt\iftrue}
\makeatother

\usepackage{xcolor} % 导入xcolor包

\usepackage{listings}

\usepackage{graphicx}
\usepackage{booktabs}
\usepackage{algorithm}
\usepackage[accsupp]{axessibility}  % Improves PDF readability for those with disabilities.

\newcolumntype{I}{!{\vrule width 0.7pt}} 
\usepackage[pagebackref,breaklinks,colorlinks,citecolor=eccvblue]{hyperref}
\usepackage{hyperref}

\usepackage{orcidlink}

\begin{document}

% ---------------------------------------------------------------
% TODO REVIEW: Replace with your title
\title{GRA: Detecting Oriented Objects through Group-wise Rotating and Attention} 

% TODO REVIEW: If the paper title is too long for the running head, you can set
% an abbreviated paper title here. If not, comment out.
\titlerunning{GRA}

% TODO FINAL: Replace with your author list. 
% Include the authors' OCRID for the camera-ready version, if at all possible.
% \author{Jiangshan Wang\inst{1}\orcidlink{0000-1111-2222-3333} \and
% Second Author\inst{2,3}\orcidlink{1111-2222-3333-4444} \and
% Third Author\inst{3}\orcidlink{2222--3333-4444-5555}}

\author{Jiangshan Wang\inst{1}$^*$ \and
Yifan Pu\inst{1}$^*$ \and
Yizeng Han\inst{2} \and Jiayi Guo\inst{1} \and \\ Yiru Wang\inst{3}\and Xiu Li\inst{1} \textsuperscript{\Letter} \and Gao Huang\inst{1} \textsuperscript{\Letter}}

% TODO FINAL: Replace with an abbreviated list of authors.
\authorrunning{J. Wang et al.}
% First names are abbreviated in the running head.
% If there are more than two authors, 'et al.' is used.

% TODO FINAL: Replace with your institution list.
\institute{Tsinghua University \and DAMO Academy, Alibaba group \and ModelTC \\
% \email{lncs@springer.com}\\
% \url{http://www.springer.com/gp/computer-science/lncs} \and
% ABC Institute, Rupert-Karls-University Heidelberg, Heidelberg, Germany\\
\email{{\{wjs23, puyf23\}@mails.tsinghua.edu.cn}, {li.xiu@sz.tsinghua.edu.cn},gaohuang@tsinghua.edu.cn}\blfootnote{$*$ Equal contribution. \Letter~Corresponding author.}}

\maketitle
\begin{abstract}
Oriented object detection, an emerging task in recent years, aims to identify and locate objects across varied orientations. This requires the detector to accurately capture the orientation information, which varies significantly within and across images. Despite the existing substantial efforts, simultaneously ensuring model effectiveness and parameter efficiency remains challenging in this scenario. In this paper, we propose a lightweight yet effective \textbf{G}roup-wise \textbf{R}otating and \textbf{A}ttention (GRA) module to replace the convolution operations in backbone networks for oriented object detection. GRA can adaptively capture fine-grained features of objects with diverse orientations, comprising two key components: Group-wise Rotating and Group-wise Attention. Group-wise Rotating first divides the convolution kernel into groups, where each group extracts different object features by rotating at a specific angle according to the object orientation. Subsequently, Group-wise Attention is employed to adaptively enhance the object-related regions in the feature. The collaborative effort of these components enables GRA to effectively capture the various orientation information while maintaining parameter efficiency. Extensive experimental results demonstrate the superiority of our method. For example, GRA achieves a new state-of-the-art (SOTA) on the DOTA-v2.0 benchmark, while saving the parameters by nearly 50\% compared to the previous SOTA method. Code is available at \url{https://github.com/wangjiangshan0725/GRA}.
% % that compared with the recent State-of-the-Art (SOTA) Adaptive Rotated Convolution method, our method not only achieves better performance (57.95\% mAP on the DOTA-v2.0 dataset) but also saves nearly 50\% network parameters.
% % Within the Group-wise Rotating, convolution kernels are divided into groups, where each group rotates an object-adaptive angle
% % with each group rotating independently at different angles, which hardly introduces additional parameters into the network. Subsequently, Group-wise Attention is employed to adaptively focus on the important regions in the feature. The collaborative effort of these components enables GRA to successfully capture the various orientation information, while maintaining the parameter efficiency. Extensive experimental results demonstrate that compared with the recent State-of-the-Art (SOTA) Adaptive Rotated Convolution method, our method not only achieves better performance (achieving the SOTA mAP of 57.95\% on DOTA-v2.0 dataset) but saves nearly 50\% network parameters. The source code will be released.
% %   \keywords{Oriented Object Detection \and Group-wise Rotating \and Group-wise Attention}
\end{abstract}

% % Despite extensive efforts, represented by the rotated kernel, have been made to address this challenge, excessive parameters are introduced into the backbone. Meanwhile, the features extracted by rotated kernels contain inaccurate regions, leading to sub-optimal detection performance.

\vspace{-0.1 in}
\section{Introduction}
\label{sec:intro}

\definecolor{blue}{RGB}{103,138,201}
\definecolor{yellow}{RGB}{191,144,0}
\definecolor{orange}{RGB}{236,130,59}
% gt里面difference大的地方也有很多物体
% ARC中对这些物体有很多没有检测到，说明旋转卷积在检测与其旋转角度不同的物体方面有缺陷，这些物体的特征里面包含了很多噪声。
% 

% 1、说明一个卷积核旋转了一个角度之后，只适合检测图像中对应角度的物体，不适合检测图像中其他角度的物体：找一张图像、一个卷积核，只检测出了对应角度的物体，其他物体没检测出来。
% 2、尽管ARC中有多个卷积核，不同卷积核的特征直接相加会导致相互干扰：用跟上面一样的图像，四个卷积核，尽管有一些不同角度的物体被检测出来了，本来那些物体检测的效果变差。
% 3、我们的方法可以做到很好的平衡
\begin{figure}[t]
	\centering
	\includegraphics[width=1.0 \textwidth]{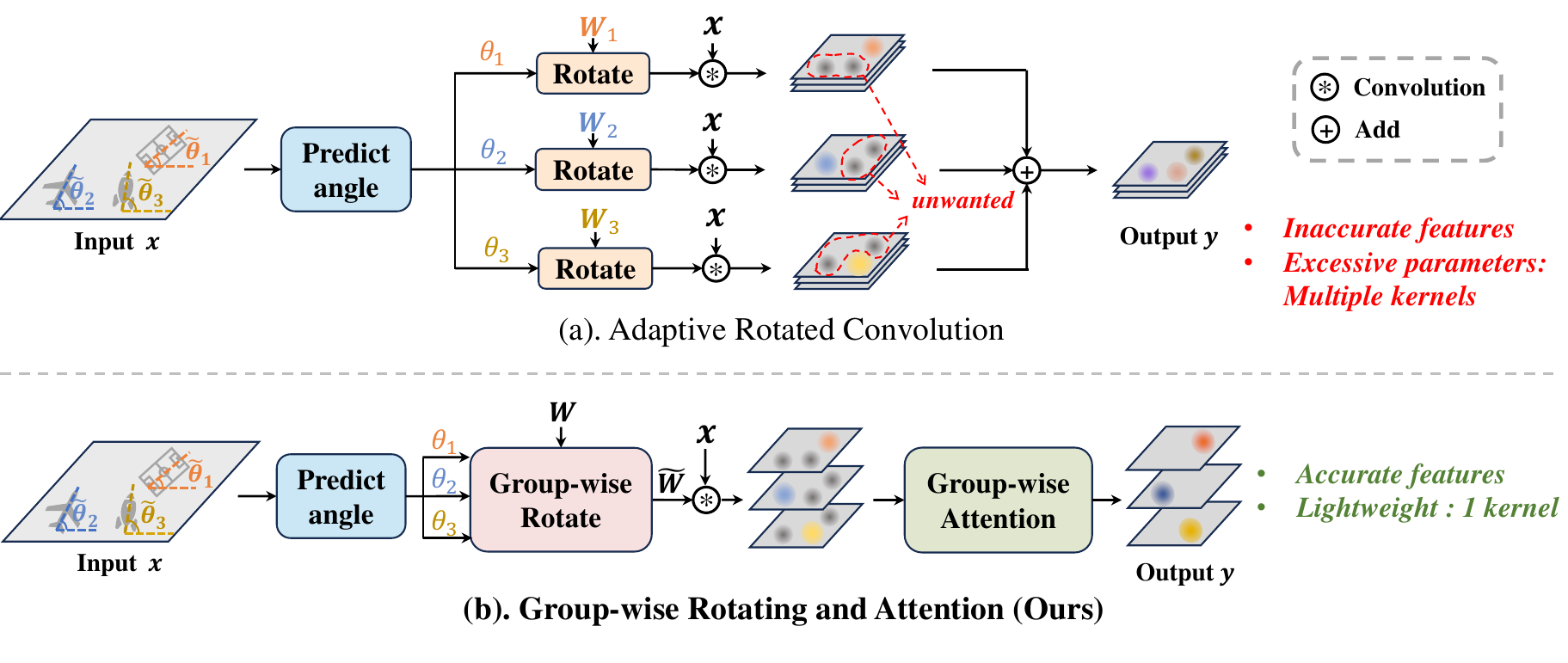} 
  % \vskip -0.15in
	\caption{\textbf{Comparison between the recent SOTA method ARC~\cite{pu2023adaptive} and our proposed method GRA.} Our method not only reduces the noise in the features but is also much more lightweight. }
	\label{fig:compare}
 % \vskip -0.31in
\end{figure}

Oriented object detection is the task of identifying and locating multiple objects with oriented bounding boxes. Different from conventional object detection which utilizes fixed horizontal bounding boxes, oriented object detection tackles a more intricate challenge, aligning better with real-world scenarios where objects frequently present various orientations. The ability to discern such oriented objects can greatly enhance situational awareness and decision-making processes across numerous industries, thereby having wide-ranging applications in image segmentation~\cite{cheng2021mask2former, yuan2020object}, autonomous driving~\cite{li2022bevformer,hu2023planning}, text recognition~\cite{yao2012detecting, karatzas2013icdar}, face detection~\cite{yang2016wider, jain2010fddb}, remote sensing~\cite{heitz2008learning, xia2018dota}, exemplar-based generation~\cite{xiao2023r, wang2024cove, guo2022assessing, ma2023follow, guo2023faceclip} and Embodied AI~\cite{xie2023learning, lv2023learning, yan2022adaptive}, \etc.

% The objects in natural images commonly exhibit various orientations, necessitating the precise detection of both their positions and orientations. Consequently, compared to conventional object detection which detects the objects with horizontal bounding box, detecting oriented objects is a more challenging task.
% The ability to discern such oriented objects can greatly enhance situational awareness and decision-making processes across numerous industries, thereby having wide-ranging applications in image segmentation~\cite{cheng2021mask2former, yuan2020object}, autonomous driving~\cite{li2022bevformer,hu2023planning}, text recognition~\cite{yao2012detecting, karatzas2013icdar}, face detection~\cite{yang2016wider, jain2010fddb}, remote sensing~\cite{heitz2008learning, xia2018dota} and Embodied AI~\cite{xie2023learning, lv2023learning, yan2022adaptive}, \etc.

Extensive efforts have been made to enhance the performance of oriented object detection from various perspectives, including the oriented bounding box representation~\cite{xu2020gliding, guo2021beyond, li2022oriented, hou2022g, yang2022detecting}, loss functions~\cite{qian2021learning, chen2020piou, yang2021rethinking, yang2021learning, yang2022kfiou, yu2024boundary}, network architecture~\cite{yang2019scrdet, yang2021r3det, yang2022scrdet++, han2021align, hou2022shape}, and label assignment strategies~\cite{ming2021sparse, ming2021dynamic}.
Recent studies~\cite{xu2023dynamic, han2021redet} indicate that the key to identifying oriented objects lies in accurately extracting their orientation information. This inspires researchers to start developing object-wise orientation-aware detection backbones. 

To this end, the recent SOTA method~\cite{pu2023adaptive} first proposes an Adaptive Rotated Convolution (ARC) module~\cite{pu2023adaptive} to replace the convolution operations in existing detection backbones. ARC employs a set of $m$ convolutional kernels, each maintaining a specific rotation angle, to separately extract the orientation information of different objects. Then the features extracted by each kernel are aggregated through a weighted sum as the output of ARC. Despite the performance improvement, ARC still encounters two primary challenges. Firstly, the utilization of $m$ convolutional kernels results in $(m-1)$ times more parameters, significantly constraining its deployment on resource-constrained devices such as remote sensing equipment. Secondly, we argue that the straightforward output feature aggregation would diminish the representation capacity of the orientation information, resulting in inaccurate detection. Specifically, we discover that a specific convolution kernel (e.g. $\textcolor{orange}{\boldsymbol{W_1}}$ in \cref{fig:compare}\textcolor{red}{a} ) primarily captures the features of objects aligned with its rotational angles (e.g. the \textcolor{orange}{ \textbf{basketball court}} in \cref{fig:compare}\textcolor{red}{a}). Consequently, features of objects at other orientations (e.g. the \textcolor{blue}{ \textbf{plane}} and the \textcolor{yellow}{ \textbf{rocket}} in \cref{fig:compare}\textcolor{red}{a}) processed by this kernel would be marred by undesired noises. 
The weighted sum of these features will mix the satisfying features and undesired noises, leading to imprecise final predictions.

To overcome these challenges, we propose the lightweight yet efficient \textbf{G}roup-wise \textbf{R}otating and \textbf{A}ttention (GRA) module (\cref{fig:compare}\textcolor{red}{b}) to capture more fine-grained orientation information. In particular, the GRA module comprises two key components: Group-wise Rotating and Group-wise Attention. The Group-wise Rotating component introduces a novel mechanism for rotation at the group level, which substantially reducing the parameter count compared to the SOTA method~\cite{pu2023adaptive}. Specifically, the convolution kernel $\boldsymbol{W} \in \mathbb{R}^{C_{\text{out}}\times C_{\text{in}}\times k\times k}$ is divided into $n$ groups along the $C_{\text{out}}$ channel. Each group of kernels is adaptively rotated based on the input feature before the convolution operation. Subsequent to Group-wise Rotating, Group-wise Attention partitions the output feature into $n$ groups similarly and applies the spatial attention mechanism. This process enables the adaptive enhancement of desirable feature regions while mitigating extraneous noise. The collaborative endeavor of these components enables extracting refined and accurate orientation information without significantly increasing the parameters of the network.

% Addressing these constraints, we propose \textbf{G}roup-wise \textbf{R}otating and \textbf{A}ttention (GRA) module (\cref{fig:compare}\textcolor{red}{b}), which comprises two components: Group-wise Rotating and Group-wise Attention. Group-wise Rotating introduces a novel group-wise rotation mechanism that significantly reduces the parameter count compared with the SOTA method~\cite{pu2023adaptive}. Specifically, the convolution kernel $\boldsymbol{W} \in \mathbb{R}^{C_{\text{out}}\times C_{\text{in}}\times k\times k}$ is divided into $n$ groups along the $C_{\text{out}}$ channel. Each group of kernels is adaptively rotated based on the input feature before the convolution operation. Subsequent to Group-wise Rotating, Group-wise Attention partitions the output feature into $n$ groups similarly and applies the spatial attention mechanism. This process enables the adaptive enhancement of desirable feature regions while mitigating extraneous noise. The collaborative endeavor of these components enables extracting the refined and accurate orientation information from images without significantly increasing the parameters of network.

The proposed GRA module is lightweight and flexible, which is
able to be seamlessly integrated into any convolution neural network to achieve improved performance. Extensive experiments on oriented object detection benchmarks including DOTA-v1.0~\cite{xia2018dota}, DOTA-v2.0~\cite{ding2021object} and HRSC2016~\cite{liu2016ship} demonstrate that GRA not only achieves a significant reduction (near 50\%) in the number of parameters compared with SOTA method \cite{pu2023adaptive} but also yields better detection performances, achieving a new SOTA result on DOTA-v2.0 dataset.

% The proposed GRA module is flexible and theoretically can be seamlessly integrated into any convolution neural network to achieve improved performance. Extensive experiments on oriented object detection benchmarks including DOTA-v1.0~\cite{xia2018dota}, DOTA-v2.0~\cite{ding2021object} and HRSC2016~\cite{liu2016ship} demonstrate that GRA not only achieves a significant reduction (near 50\%) in the number of parameters compared with SOTA method \cite{pu2023adaptive} but also yields better detection performances, achieving a new SOTA result on DOTA-v2.0 dataset.

\section{Related work}

\indent
{\bf Oriented Object Detection.} 
Oriented object detection advances the conventional horizontal object detection paradigm by employing oriented bounding boxes, making it a more precise task. 
Numerous studies have been dedicated to devising specialized detectors for rotated objects, including various components such as enhancing features in the detector neck~\cite{yang2019scrdet, yang2021r3det, yang2022scrdet++}, the oriented region proposal network~\cite{xie2021oriented, cheng2022anchor}, the mechanism for extracting rotated regions of interest (RoI)~\cite{ding2019learning, xie2021oriented}, the design of the detector head~\cite{han2021align, hou2022shape}, and the utilization of advanced label assignment strategies~\cite{ming2021sparse, ming2021dynamic}.
Another research direction~\cite{li2022oriented, xu2020gliding, guo2021beyond, hou2022g} focuses on developing more adaptable representations of objects. Simultaneously, there has been extensive research into devising suitable loss functions for various oriented object representations~\cite{yang2021rethinking, yang2021learning, yang2022kfiou}. Besides oriented object detection, there are also some works focusing on detecting other kinds of objects \cite{he2023camouflaged,he2024strategic,he2024weakly} or even video frames \cite{xiao2023bridging}, which are more challenging than conventional object detection.

{\bf Design of Backbone for Oriented Object Detection.} 
Recent endeavors in the field of oriented object detection have increasingly concentrated on optimizing the architecture of backbone networks for oriented object detection. ReDet~\cite{han2021redet} integrates rotation-equivariant operations within the backbone, yielding features that maintain orientation fidelity. However, such operations may not fully account for the diversity of object orientations within individual images or across datasets. LSKNet~\cite{li2023large} proposes to use spatial attention~\cite{woo2018cbam, xia2023dat++} to adaptively choose a suitable kernel size for the input image while the orientation information is also not taken into account. ARC~\cite{pu2023adaptive} proposes to extract the orientation information of oriented objects by dynamically rotating the convolution kernels conditioned on different image samples. On the other hand, a large number of additional parameters are introduced and the features of different angles are mixed, introducing the undesired noises.

{\bf Dynamic Neural Networks.} 
Contrary to static models, which maintain constant computational graphs and parameters throughout inference, dynamic neural networks~\cite{han2021dynamic,wang2023computation, pu2024mediators} can modify their architecture or parameters in response to varying inputs.
Dynamic networks are typically categorized into three distinct types: sample-wise~\cite{huang2017multi,wang2021not,han2022learning, han2023dynamic, pu2023fine, guo2023zero,xiao2024grootvl, wang2023erratum}, spatial-wise~\cite{huang2022glance,han2022latency,han2023latency,han2021spatially}, and temporal-wise~\cite{hansen2019neural,wang2021adaptive, guo2023smooth}. Recently, with the development of query-based Detection Transformers~\cite{hu2024dac, zhao2024ms, shen2023v}, a new kind of query-based dynamic network has begun to flourish~\cite{pu2023rank}.
In this work, we introduce a sample-wise dynamic network, which dynamically adjusts the model parameters conditioned on different input images. To be specific, the convolution kernels are rotated in a group-wise manner, which endows convolution parameters with a better ability to capture orientation information in the images.

\section{Method}

\definecolor{OliveGreen}{rgb}{0.35, 0.60, 0.24}

\begin{figure}[t]
	\centering
	\includegraphics[width=1.0 \textwidth]{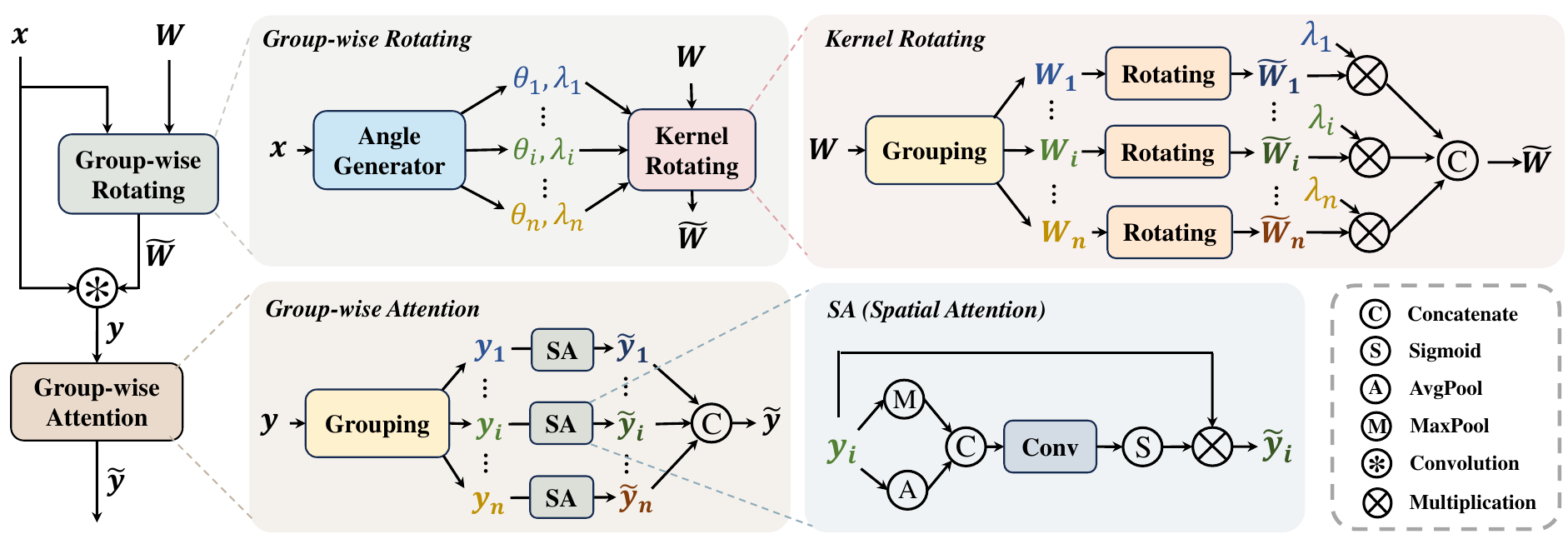} 
	\caption{\textbf{An overview of our proposed GRA method}, which contains two components: Group-wise Rotating and Group-wise Attention. In Group-wise Rotating module, the input kernel $\boldsymbol{W} \in \mathbb{R}^{C_{\text{out}}\times C_{\text{in}}\times k\times k}$ is rotated in a group-wise manner, obtaining $\boldsymbol{\widetilde{W}}$, which performs the convolution with input feature map $\boldsymbol{x}\in \mathbb{R}^{C_{\text{in}}\times H_{\text{in}} \times W_{\text{in}}}$. The output $\boldsymbol{y}\in \mathbb{R}^{C_{\text{out}}\times H_{\text{out}} \times W_{\text{out}}}$ is then fed into the group-wise attention module for denoising and refining, obtaining the final output $\boldsymbol{\widetilde{y}}$. }
	\label{fig:pipeline}
 % \vskip -0.2in
\end{figure}

In this section, we firstly introduce the current SOTA method~\cite{pu2023adaptive} and analyze its limitations. Subsequently, we delineate our proposed GRA module, comprising Group-wise Rotating and Group-wise Attention. Within the Group-wise Rotating module, a lightweight network named angle generator is employed to predict $n$ rotation angles from the input feature map ${\boldsymbol{x}}$. Then the convolution kernel $\boldsymbol{W} \in \mathbb{R}^{C_{\text{out}}\times C_{\text{in}}\times k\times k}$ is partitioned into $n$ groups at $C_{\text{out}}$ dimension, each rotating to the corresponding angle predicted by the angle generator. These rotated groups of kernels are subsequently concatenated together to perform the convolution with ${\boldsymbol{x}}$, obtaining the output feature ${\boldsymbol{y}}$. Followed by Group-wise Rotating, the Group-wise Attention module is proposed to adaptively emphasize satisfying regions of each feature group in ${\boldsymbol{y}}$ while attenuating extraneous noise. The final output of GRA module ${\boldsymbol{\widetilde{y}}}$ is fed into subsequent modules in the network. Compared with the recent SOTA method~\cite{pu2023adaptive}, GRA achieves a substantial reduction of parameters and effectively mitigates the issue of inaccurate features in an elegant and reasonable manner. Finally, we will discuss the remarkable advantages of GRA. An overview of our method is shown in \cref{fig:pipeline}. 

\subsection{Preliminary}
\label{sec:pre}
There have been several previous works focusing on developing specialized backbones for oriented object detection. Among them, Adaptive Rotated Convolution (ARC) \cite{pu2023adaptive} proposes to dynamically adjust the orientation of convolutional kernels based on the objects present within the image. This method improves the extraction of features from oriented objects, thereby achieving state-of-the-art (SOTA) performance across various benchmarks. In an ARC module, $m$ convolution kernels $\{\boldsymbol{W}_i\}$, where $\boldsymbol{W}_i \in \mathbb{R}^{C_{\text{out}}\times C_{\text{in}}\times k\times k}$ and ${i \in \{1,2,\cdots,m\}}$, are applied to capture features of multiple oriented objects. A routing function is employed which takes the feature map $\boldsymbol{x}$ as the input, predicting the angle $\theta _i$ and the weight $\lambda _i $ for each kernel $\boldsymbol{W}_i$. 
\begin{equation}
\{\theta _i\}_{i\in \{1,2,...,m\}}, \{\lambda _i\}_{i\in \{1,2,...,m\}} = \text{Routing}(\boldsymbol{x}),
\end{equation}
Subsequently, the kernels are rotated at the corresponding predicted angles, obtaining $\{\boldsymbol{\widetilde{W}}_i\}$, where each $\boldsymbol{\widetilde{W}}_i = \text{Rotate}(\boldsymbol{W}_i, \theta_i)$ and ${i \in \{1,2,\cdots,m\}}$. Each $\boldsymbol{\widetilde{W}}_i$ independently performs convolution on the input feature map $\boldsymbol{x}$, yielding $\boldsymbol{y}_i$. These intermediate results are then aggregated through a weighted sum to produce the final output $\boldsymbol{y}$,
{
\setlength{\abovedisplayskip}{5pt}
\setlength{\belowdisplayskip}{5pt}
\begin{gather}
 \ \ \ 
\boldsymbol{y}_i = \text{Conv}(\boldsymbol{x}, \boldsymbol{\widetilde{W}}_i),\ \ \ 
\boldsymbol{y} = \sum_{i=1}^m \lambda_i \times \boldsymbol{y}_i.
\end{gather}
}

The resultant feature $\boldsymbol{y}$ is fed into the following modules. In experiments, the value of $m$ is generally assigned to a relatively small value (typically set to 4).

\subsection{Limitations Analysis}
\label{sec:lim}

While ARC~\cite{pu2023adaptive} exhibits superior performance compared to traditional convolution, there still exist two main shortcomings as illustrated following:  

\textbf{Excessive parameters}. Within each ARC module, $m$ convolution kernels $\boldsymbol{W}_i$ are employed, multiplying the number of parameters by $m$ compared to a standard convolution layer. The escalation in parameters is proportional to the factor $m$, thereby amplifying the complexity of the network. For instance, a standard ResNet50 comprises approximately 23.5M parameters; However, integrating ARC with $m=4$ inflates the total parameters to the overwhelming 57.2M. Such expansion poses a considerable challenge for deployment in resource-limited settings such as remote sensing devices, where storage space is at a premium and cannot support models with extensive parameters.

\begin{figure}[t]
	\centering
	\includegraphics[width=1.0 \textwidth]{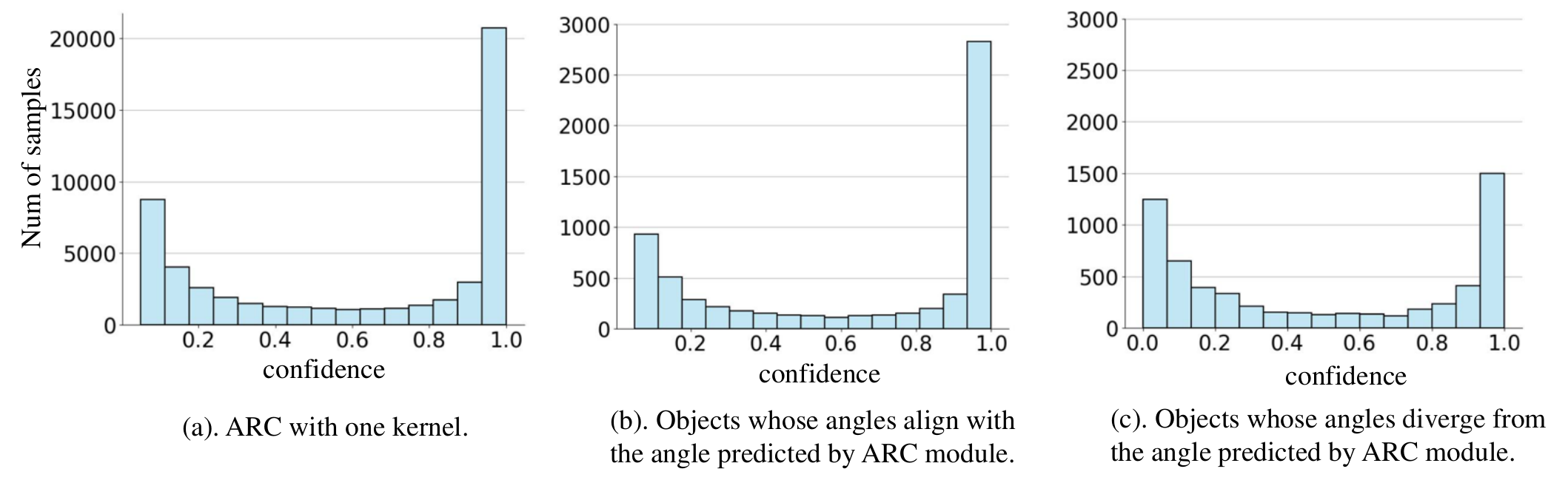} 
 % \vskip -0.1in
	\caption{\textbf{Distribution of the confidence predicted by ARC ($m=1$)}. The angle predicted by ARC module can affect the final prediction of the objects with different angles in the images. In general, the objects whose orientation is close to the angle predicted by ARC module can be detected with higher confidence compared to the objects whose orientation diverges from the ARC-predicted angle.}
	\label{fig:mot_arc}
 % \vskip -0.10in
\end{figure}

\begin{figure}[t]
	\centering
	\includegraphics[width=1.0 \textwidth]{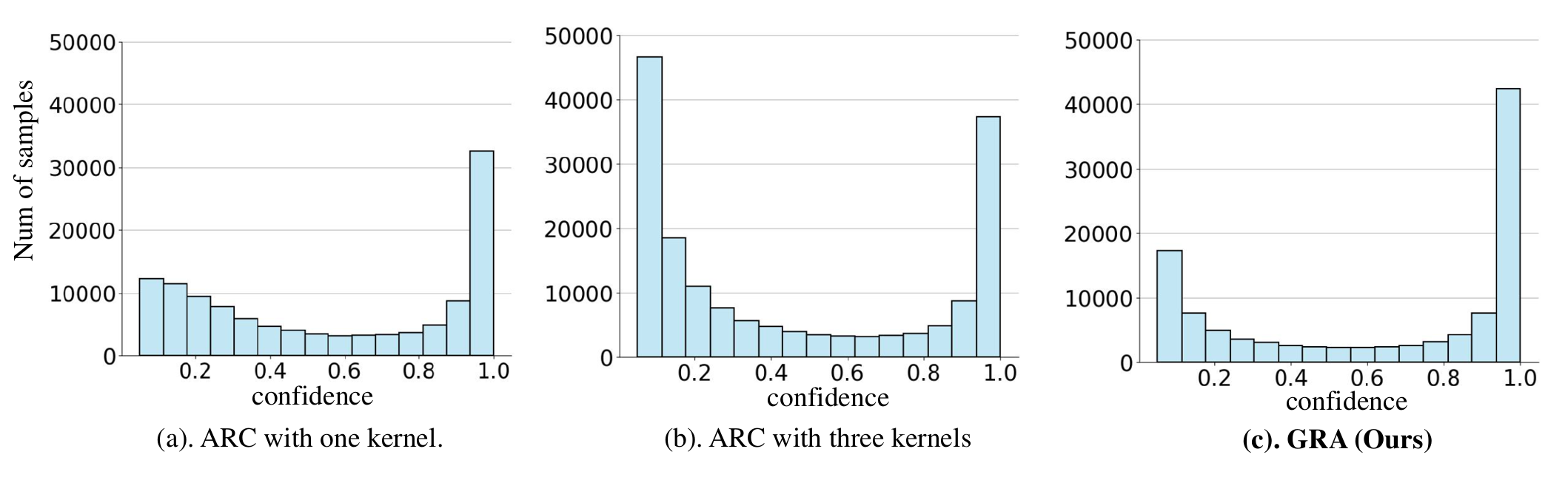} 
 % \vskip -0.10in
	\caption{\textbf{Comparison of the prediction between ARC and our method.} The weighted sum of ARC can lead to inaccurate features, causing a number of low-confidence detections. On the other hand, our method can detect more objects in the images with higher confidence.}
	\label{fig:mot_comp}
 % \vskip -0.2in
\end{figure}

\textbf{Inaccurate features}. Within the ARC framework, features obtained from convolution kernels rotated at various angles are aggregated through a weighted sum. While this mechanism aims to detect objects at different rotational orientations within images, it introduces inaccurate information and reduces the representational capacity of the output feature $\boldsymbol{y}$. To illustrate this issue, we train an Oriented RCNN \cite{xie2021oriented} with the ARC module (setting $m=1$) on the training set from DOTA. We then evaluate the performance of the detector on the validation set and visualize the distribution of confidences across all detected objects (\cref{fig:mot_arc}\textcolor{red}{a}). To examine the influence of the ARC module on the detector's predictions, we specifically consider objects whose ground truth orientations closely align with the angle predicted by the ARC module (i.e. the angle for rotating the kernel), analyzing the confidence of these objects predicted by the detector. For these objects, the detector detects them with a predominant majority demonstrating high confidence levels as depicted in \cref{fig:mot_arc}\textcolor{red}{b}. Conversely, for objects whose ground truth orientations diverge from the angle predicted by ARC module, the detector struggles to accurately detect them, resulting in numerous low-confidence detections (\cref{fig:mot_arc}\textcolor{red}{c}). This indicates that the features extracted by the backbone for these objects are inadequate (corresponds to the unwanted features in \cref{fig:compare}), thereby hindering the subsequent detector head from accurately identifying these objects. Additional experiments conducted on the test set of DOTA (as shown in Figure \ref{fig:mot_comp}) further elucidate the impact of integrating the ARC and GRA modules respectively. Integrating ARC with three kernels ($m=3$) into the backbone enables the detector to identify a greater number of oriented objects in images, consequently leading to improved performance compared to using ARC with just one kernel ($m=1$). However, it also results in a notable increase in low-confidence detections (\cref{fig:mot_comp}\textcolor{red}{b}), some of which should have been detected with high confidence. This observation suggests that a direct weighted sum of features is sub-optimal, potentially compromising the features obtained by various rotated kernels. In contrast, our approach mitigates this issue effectively, resulting in significantly improved performance (\cref{fig:mot_comp}\textcolor{red}{c}).

\subsection{Group-wise Rotation}
In this section, we will comprehensively introduce the Group-wise Rotation module within GRA, which comprises three integral components: Predicting Rotating Angle, Grouping, and Rotating.

\textbf{Predicting Rotating Angle.} To predict the rotation angle corresponding to objects within a feature map $\boldsymbol{x}\in \mathbb{R}^{C_{\text{in}}\times H_{\text{in}}\times W_{\text{in}}}$, we employ a lightweight network named angle generator. 
Initially, the feature map $\boldsymbol{x}$ undergoes processing through a depth-wise convolution layer, followed by ReLU activation and LayerNorm layers, which aim to extract inherent orientation information from diverse objects. Subsequently, a global pooling layer transforms the feature into a vector with dimension $C_{\text{in}}$. This resulting vector then passes through two distinct linear layers, each with $n$ output channels, producing a series of predicted angles $\{\theta_j\}_{j\in \{1, 2, \cdots, n\}}$ and scale factors $\{\lambda_j\}_{j\in \{1, 2, \cdots, n\}}$.

These predicted angles and scale factors are utilized for the group-wise rotation of convolution kernels. Notably, the number of angles $n$ is significantly smaller than the output channels of convolution, $C_{\text{out}}$. Consequently, it is convenient to partition the convolution kernels into $n$ distinct groups at the $C_{\text{out}}$ channel, with each group subjected to a unique rotation angle.

\textbf{Grouping.} The convolution operation utilizes a kernel $\boldsymbol{W} \in \mathbb{R}^{C_{\text{out}}\times C_{\text{in}}\times k\times k}$ which takes $\boldsymbol{x}\in \mathbb{R}^{C_{\text{in}}\times H_{\text{in}}\times  W_{\text{in}}}$ as the input to produce the output feature $\boldsymbol{y}\in \mathbb{R}^{C_{\text{out}}\times H_{\text{out}}\times  W_{\text{out}}}$. This operation involves $C_{\text{out}}$ individual kernels $\boldsymbol{w}_i \in \mathbb{R}^{C_{\text{in}}\times k\times k}$, $i \in \{1,2,\cdots,C_{\text{out}}\}$, convolving with $\boldsymbol{x}$ separately. The resulting outputs are then concatenated to obtain the final output $\boldsymbol{y}$. In our method, these $C_{\text{out}}$ kernels are uniformly divided into $n$ groups $\{\boldsymbol{W}_j$\},
\begin{align}
\boldsymbol{W} &=\{\boldsymbol{w}_i\},  {i\in\{1,2,\cdots,C_{\text{out}}\}} \\
&=\{\boldsymbol{W}_j\}, {j\in\{1,2,\cdots,n\}}.
\end{align}
Each group $\boldsymbol{W}_j$ contains $C_{\text{out}}/n$ kernels $\boldsymbol{w}_{j,l}$,
\begin{equation}
\boldsymbol{W}_j=\{\boldsymbol{w}_{j,l}\}, l\in \{1,2,\cdots,C_{\text{out}}/n\}.  \\
% \boldsymbol{k}_{j,l}=\boldsymbol{k}_{l+{(j-1)\times C_{\text{out}}/n}}.
\end{equation}
By employing the grouping strategy, kernels in different groups work independently for the ensuing rotation operation. 

\textbf{Rotating.} After predicting rotational angles and grouping kernels, each kernel within the $j$th group $\boldsymbol{W}_j$ undergoes rotation by the corresponding angle $\theta _j$, resulting in the rotated kernel group $\boldsymbol{{\widetilde{W}}}_j$. Additionally, each group of kernels is scaled by the corresponding learnable scale factor $\lambda_j$, which represents the relative importance of the different groups.
\begin{equation}
\boldsymbol{{\widetilde{W}}}_j=\{\lambda_j \times  \text{Rotate}(\boldsymbol{w}_{j,l},\theta _j)\}, l\in \{1,2,...,C_{\text{out}}/n\}.
\end{equation}

The rotation process is executed using bilinear interpolation as described in \cite{pu2023adaptive}. Specifically, considering a $k\times k$ kernel, the value at each point within this kernel can be seen as the value of a sampled point from the kernel space. During rotating, the values of the original $k\times k$ kernel are used to span the entire kernel space using bi-linear interpolation. Subsequently, we calculate the new position of each point after rotating by $\theta$ and obtain its value in the kernel space. The final rotated kernel is obtained by aggregating the different groups of kernels, i.e, $\boldsymbol{\widetilde{W}}=\{\boldsymbol{\widetilde{W}}_{j}\}$, ${j\in\{1,2,...,n\}}$. The implementation of group-wise rotating is simple and efficient, accomplished through a single step of matrix multiplication to transform $\boldsymbol{W}$ into $\boldsymbol{{\widetilde{W}}}$.

Followed Group-wise Rotating, the rotated kernel $\boldsymbol{{\widetilde{W}}}$ is utilized to perform convolution with the input feature map $\boldsymbol{x}$, i.e., $\boldsymbol{y} = \text{Conv}(\boldsymbol{x}, \boldsymbol{{\widetilde{W}}})$. The resulting output $\boldsymbol{y} \in \mathbb{R}^{C_{\text{out}}\times H_{\text{out}}\times W_{\text{out}}}$ is then passed into the Group-wise Attention module for further processing.

\textbf{Differences between Group Convolution. }Although kernels are divided into different groups in our method, it is distinct from Group Convolution. In Group Convolution, the input feature map $\boldsymbol{x} \in \mathbb{R}^{C_{\text{in}}\times H_{\text{in}}\times W_{\text{in}}}$ is partitioned into $g$ groups along the $C_{\text{in}}$ dimension. Each group then undergoes convolution with a corresponding set of ${C_{\text{out}}}/{g}$ kernels, each with dimensions $\mathbb{R}^{{C_{\text{in}}}/{g}\times k\times k}$. In contrast, our grouping strategy is employed to rotate the $C_{\text{out}}$ kernels. After the rotation, regular convolution is applied to obtain the output.

\subsection{Group-wise Attention}

Following the convolution operation between $\boldsymbol{x}$ and $\boldsymbol{{\widetilde{W}}}$, the resulting feature $\boldsymbol{y}$ naturally comprises $n$ groups, i.e. $\boldsymbol{y}=\{\boldsymbol{y}_j\}, {j\in\{1,2,...,n\}}$. These feature groups $\boldsymbol{y}_j\in \mathbb{R}^{C_{\text{out}}/n \times H_{\text{out}}\times W_{\text{out}}}$ predominantly capture the relevant characteristics of objects oriented near $\theta_j$. However, features corresponding to objects at other angles may contain undesired noise and are sub-optimal. To mitigate this issue, we introduce a group-wise spatial attention mechanism to selectively amplify important regions within each feature group while suppressing the influence of irrelevant areas. To be specific, max and average pooling operations are conducted on each feature group $\boldsymbol{y}_j$, and their outputs are concatenated, 
\begin{equation}
\boldsymbol{S}_j = \text{Concat}[\text{AvgPool}(\boldsymbol{y}_j), \text{MaxPool}(\boldsymbol{y}_j)]. \\
\end{equation}
This operation highlights the most important regions within each feature group while preserving the overall representation of average characteristics. The pooled feature $\boldsymbol{S}_j \in \mathbb{R}^{2 \times H_{\text{out}}\times W_{\text{out}}}$ is passed through a convolution layer $F$ for channel adjustment. Subsequently, a Sigmoid function $\sigma$ is applied to map all values to the range [0,1], yielding the attention maps: 
$\widetilde{\boldsymbol{S}}_j = \sigma(F(\boldsymbol{S}_j))$,  $\widetilde{\boldsymbol{S}}_j \in \mathbb{R}^{1 \times H_{\text{out}}\times W_{\text{out}}} $.

The final output feature $\widetilde{\boldsymbol{y}}_j$ is obtained through element-wise multiplication between $\boldsymbol{y}_j$ and $\widetilde{\boldsymbol{S}}_j$, i.e., 
$\widetilde{\boldsymbol{y}}_j = \boldsymbol{y}_j \odot \widetilde{\boldsymbol{S}}_j$.   
This group-wise attention mechanism enhances the important regions within the feature, simultaneously attenuating less relevant areas. The resulting final output $ \boldsymbol{\widetilde{y}}=\{\boldsymbol{{\widetilde{y}}}_j\}$, where ${j\in\{1,2,...,n\}}$, is then forwarded to subsequent modules in the network for further processing.

\subsection{Discussion}
\textbf{Enhanced Detection of Fine-Grained Orientation Details.} Compared with ARC, the proposed grouping mechanism enables the angle generator to predict more angles (with $n$ set to 16 in our experiments) from the input feature map while introducing minimal extra parameters. This strategy enables a more comprehensive capture of subtle orientation nuances within the input feature map. Therefore, the leverage of the group-wise rotated kernel for convolution enhances the learning of object features across diverse orientations. Moreover, the Group-wise Attention further refines these features, enhancing the precision and quality of the extracted information.

\textbf{Relationship between Group-wise Rotating and Group-wise Attention.} The synergy between Group-wise Rotating and Group-wise Attention within the GRA module is essential and irreplaceable. The Group-wise Rotating mechanism, by dynamically adjusting the convolutional kernel to different orientations, enables the network to effectively capture object features across a range of angles. However, as discussed in Section \ref{sec:lim}, this process inadvertently introduces undesired noise into the features. To address this issue, Group-wise Attention is ingeniously employed to adaptively filter out undesirable features while reinforcing the relevant regions of the feature map. Under the joint effect of these two components, our method effectively captures the satisfying features of objects with various orientations within a single image.

% 这两个部分是紧密相关、相辅相成的。
% 在group-wise rotating中，通过将卷积核旋转不同的角度，图像中不同角度物体的特征被提取出来。但是并不是所有的特征都是满意的。我们通过group-wise attention巧妙地解决了这个问题，自适应地去除了不需要的特征而保留了需要的特征

\section{Experiment}

\subsection{Experiment Setup}

\textbf{Datasets.} We evaluate our methods on three widely-used datasets in oriented object detection, i.e., DOTA-v1.0~\cite{xia2018dota}, DOTA-v2.0~\cite{ding2021object} and HRSC2016~\cite{liu2016ship}.
DOTA-v1.0~\cite{xia2018dota} contains 2806 aerial images and 188,282 objects across various categories, each annotated with precisely oriented bounding boxes. It includes 15 object classes. DOTA-v2.0~\cite{ding2021object} contains 11,268 images and 1,793,658 objects. Besides the 15 classes in the DOTA-v1.0 dataset, the DOTA-v2.0 dataset contains 3 additional classes. For these two datasets, we use the official training set and validation set for training. The testing is conducted on the test set. The mean average precision (mAP) on the test set is obtained by submitting the prediction of the model to the official DOTA dataset evaluation server.

HRSC2016~\cite{liu2016ship} is also a widely-used benchmark for arbitrary-oriented object detection. There are 1061 images with sizes ranging from 300×300 to 1500×900. In the experiment, both the training set (436 images) and the validation set (181 images) are utilized for training. The test set is used for testing. We report the COCO~\cite{lin2014microsoft} style mean average precision (mAP) on the test set to illustrate the effectiveness of our model. During data pre-processing, we maintain the original aspect ratios of images.

\textbf{Implementation Details.}
GRA is a plug-and-play module and theoretically can be inserted into any convolution network. In the experiment, we replace $3 \times 3$ convolutions in the last three stages of ResNet, preserving $1 \times 1$ convolutions due to their inherent rotational invariance. Unless specifically stated, the backbone is pretrained for 100 epochs on ImageNet before training on the aforementioned dataset for oriented object detection.

For the DOTA dataset, the total training epoch is set to 12 unless specifically stated. For HRSC2016 dataset, Rotated RetinaNet~\cite{lin2017focal} is trained for 72 epochs, both S$^{2}$ANet~\cite{han2021align} and Oriented R-CNN~\cite{xie2021oriented} are trained for 36 epochs. We scale down the learning rate of the backbone during training, making the training procedure of GRA more stable. 
Most of the experiments in our work are implemented with MMRotate framework \cite{zhou2022mmrotate} except the experiments on Oriented R-CNN, which are implemented with the OBBDetection framework \cite{xie2021oriented}. 
% To ensure the fairness of comparison, we maintain consistent training configurations for both our proposed method and the baseline methods across all experiments. 

\setlength\tabcolsep{4pt}
\begin{table}[ht]

  \caption{{\bf GRA outperforms the SOTA method  ARC~\cite{pu2023adaptive} on the DOTA-v1.0 dataset among various oriented object detectors}. }
  % \vskip -0.12in
  \label{tab:headings}
  \centering
  % \resizebox{1.0\textwidth}{!}{
  \resizebox{0.95\textwidth}{!}{
    \begin{tabular}{l | c | c c   l  | c}
      \toprule
      \multirow{2}{*}{Method} & \multirow{2}{*}{Backbone} & \multicolumn{3}{c|}{Params(M)} & \multirow{2}{*}{mAP(\%)} \\

      & & Backbone & \, \,\,Head\, \,\, & Total & \\
      \midrule
    
    \multirow{3}{*}{{Rotated  RetinaNet~\cite{lin2017focal}}} 
    
    &  {R50} 
    &  {23.51} 
    &  {13.13}
    &  {36.64}
    &  {68.42}\\
    
    &  {$\text{R50}_{\text{ARC}}$} 

    &  {57.20} 
    &  {13.13}
    &  {70.33}
    &  {71.45}\\
    
    & \cellcolor{lightgray!30 }{\bf $\text{R50}_{\text{GRA}}$} 
    
    & \cellcolor{lightgray!30}{23.79} 
    & \cellcolor{lightgray!30}{13.13}
    & \cellcolor{lightgray!30}{\bf 36.92}$_{\textcolor{OliveGreen}{(\downarrow \textbf{46\%})}}$
    & \cellcolor{lightgray!30}{\bf 72.19}

    \\
    %mAP: 0.5797, AP50: 0.8980, AP75: 0.6840
    \midrule
    \multirow{3}{*}{R3Det~\cite{yang2021r3det}}

    &  {R50} 
    
    &  {23.51} 
    &  {18.62}
    &  {42.13}
    &  {69.70}\\
    
    &  {$\text{R50}_{\text{ARC}}$} 
    
    &  {57.20} 
    &  {18.62}
    &  {75.82}
    &  {72.32}\\
    
    & \cellcolor{lightgray!30}{\bf $\text{R50}_{\text{GRA}}$} 
    
    & \cellcolor{lightgray!30}{23.79} 
    & \cellcolor{lightgray!30}{18.62}
    & \cellcolor{lightgray!30}{\bf 42.41}$_{\textcolor{OliveGreen}{{(\downarrow \textbf{43\%})}}}$

    & \cellcolor{lightgray!30}{\bf 72.93} 
    
    \\
    %mAP: 0.5797, AP50: 0.8980, AP75: 0.6840
    \midrule
    \multirow{3}{*}{S$^{2}$ANet~\cite{han2021align}}

    &  {R50} 
    
    &  {23.51} 
    &  {15.32}
    &  {38.83}
    &  {74.13}\\
    
    &  {$\text{R50}_{\text{ARC}}$} 
    
    &  {57.20} 
    &  {15.32}
    &  {72.52}
    &  {75.49}\\

    & \cellcolor{lightgray!30}{\bf $\text{R50}_{\text{GRA}}$} 
    
    & \cellcolor{lightgray!30}{23.79} 
    & \cellcolor{lightgray!30}{15.32}
    & \cellcolor{lightgray!30}{\bf 39.11}$_{\textcolor{OliveGreen}{{(\downarrow \textbf{45\%})}}}$
    & \cellcolor{lightgray!30}{\bf 75.98}

    \\
    %mAP: 0.5797, AP50: 0.8980, AP75: 0.6840
    \midrule
    % \multirow{2}{*}{Rotated Faster \\ RCNN~\cite{ren2015faster}}
    \multirow{3}{*}{{Rotated Faster RCNN~\cite{ren2015faster}} }
    &  {R50} 
    
    &  {23.51} 
    &  {17.86}
    &  {41.37}
    &  {73.17}\\
    
    &  {$\text{R50}_{\text{ARC}}$} 
    
    &  {57.20} 
    &  {17.86}
    &  {75.06}
    &  {74.77}\\
    
    & \cellcolor{lightgray!30}{\bf $\text{R50}_{\text{GRA}}$} 
    
    & \cellcolor{lightgray!30}{23.79} 
    & \cellcolor{lightgray!30}{17.86}
    & \cellcolor{lightgray!30}{\bf 41.65}$_{\textcolor{OliveGreen}{(\downarrow \textbf{43\%})}}$
    & \cellcolor{lightgray!30}{\bf 75.22}
    
    \\
    %mAP: 0.5797, AP50: 0.8980, AP75: 0.6840
    \midrule
    \multirow{3}{*}{CFA~\cite{guo2021beyond}}

    &  {R50} 
    
    &  {23.51} 
    &  {13.33}
    &  {36.84}
    &  {69.37}\\
    
    &  {$\text{R50}_{\text{ARC}}$} 
    
    &  {57.20} 
    &  {13.33}
    &  {70.53}
    &  {73.53}\\
    
    & \cellcolor{lightgray!30}{\bf $\text{R50}_{\text{GRA}}$} 
    
    & \cellcolor{lightgray!30}{23.79} 
    & \cellcolor{lightgray!30}{13.33}
    & \cellcolor{lightgray!30}{\bf 37.12}$_{\textcolor{OliveGreen}{(\downarrow \textbf{46\%})}}$
    & \cellcolor{lightgray!30}{\bf 73.97} 

    \\

    \midrule
    \multirow{3}{*}{{Oriented R-CNN~\cite{xie2021oriented}}} 

    &  {R50} 
    
    &  {23.51} 
    &  {17.86}
    &  {41.37}
    &  {75.81}\\
    
    &  {$\text{R50}_{\text{ARC}}$} 
    
    &  {57.20} 
    &  {17.86}
    &  {75.06}
    &  {77.35}\\
    
    & \cellcolor{lightgray!30 }{\bf $\text{R50}_{\text{GRA}}$} 
    
    & \cellcolor{lightgray!30}{23.79} 
    & \cellcolor{lightgray!30}{17.86}
    & \cellcolor{lightgray!30}{\bf 41.65}$_{\textcolor{OliveGreen}{(\downarrow \textbf{43\%})}}$ 
    & \cellcolor{lightgray!30}{\bf 77.63} 
    \\
    
    % \midrule
    % \cmidrule{2-6}
    % &  {R101} 
    
    % &  {42.50} 
    % &  {17.86}
    % &  {60.36}
    % &  {76.11}\\
    
    % &  {$\text{R101}_{\text{ARC}}$} 
    
    % &  {106.35} 
    % &  {17.86}
    % &  {124.21}
    % &  {77.70}\\
    
    %  & \cellcolor{lightgray!30 }{\bf $\text{R101}_{\text{Ours}}$} 
     
    % & \cellcolor{lightgray!30}{42.84} 
    % & \cellcolor{lightgray!30}{17.86}
    % & \cellcolor{lightgray!30}{\bf 60.70}$_{\textcolor{OliveGreen}{(\downarrow \textbf{51\%})}}$ 
    % & \cellcolor{lightgray!30}{\bf 77.85} 
    % \\
    
    \bottomrule
  \end{tabular}
  }

\label{tab:main}
% \vspace{-0.3cm}
\end{table}

\begin{table}[ht]

  \caption{{\bf Results on HRSC2016.} The proposed group-wise rotating and attention mechanism is also effective in some widely used oriented object detectors.}
  % \vspace{-0.3cm}
  \label{tab:headings}
  \centering
  \resizebox{0.8\textwidth}{!}{
  \begin{tabular}{l | c | c | c c c}
    \toprule
    Method & Backbone & Params(M) & AP$_\text{50}$(\%) & AP$_\text{75}$(\%) & mAP(\%) \\
    \midrule
    \multirow{3}{*}{{Rotated RetinaNet~\cite{lin2017focal}}} 
    
    & R50 & 36.64 & 84.20 & 58.50 & 52.70 \\
    
    &  {$\text{R50}_{\text{ARC}}$} 
    & 70.33
    &  {85.10} 
    &  {60.20}
    &  {53.97} \\
    
    & \cellcolor{lightgray!30 }{\bf $\text{R50}_{\text{GRA}}$} 
    & \cellcolor{lightgray!30}{\bf 36.92} 
    & \cellcolor{lightgray!30}{\bf 85.20} 
    & \cellcolor{lightgray!30}{\bf 60.34}
    & \cellcolor{lightgray!30}{\bf 54.08}%$_{(\textcolor{blue}{\uparrow \textbf{1.27}})}$

    \\
    %mAP: 0.5797, AP50: 0.8980, AP75: 0.6840
    \midrule
    \multirow{3}{*}{S$^{2}$ANet~\cite{han2021align}}
    & R50 & 38.83 & 89.70 & 65.30 & 55.65 \\
    &  {$\text{R50}_{\text{ARC}}$} 
    & 72.52
    &  {90.00} 
    &  {67.40}
    &  {57.77} \\
    & \cellcolor{lightgray!30}{\bf $\text{R50}_{\text{GRA}}$} 
    & \cellcolor{lightgray!30}{\bf 39.11}
    & \cellcolor{lightgray!30}{\bf {90.04}}
    & \cellcolor{lightgray!30}{\bf 68.40}
    & \cellcolor{lightgray!30}{\bf 58.03}\\%$_{(\textcolor{blue}{\uparrow \textbf{2.32}})}$ 
    
    \midrule

    \multirow{3}{*}{{Oriented R-CNN~\cite{xie2021oriented}}} 
    & R50 & 41.37 & 90.40 & 88.81 & 70.55 \\
    
    &  {$\text{R50}_{\text{ARC}}$} 
    & 75.06
    &  {90.41} 
    &  {89.02}
    &  {72.39} \\
     & \cellcolor{lightgray!30 }{\bf $\text{R50}_{\text{GRA}}$} 
    & \cellcolor{lightgray!30}{\bf 41.65} 
    & \cellcolor{lightgray!30}{\bf 90.48} 
    & \cellcolor{lightgray!30}{\bf 89.23}
    & \cellcolor{lightgray!30}{\bf 72.59}\\%$_{(\textcolor{blue}{\uparrow \textbf{1.90}})}$ 
    \bottomrule
  \end{tabular}
  }
\label{tab:hrsc}
\end{table}
\begin{table}[ht]
\centering
\begin{minipage}{0.48\textwidth}
\centering
\caption{{\bf Experiment results on the DOTA dataset under multi-scale training and testing.} Although our method cut down the parameters significantly, it still outperforms a number of SOTA methods.}
% \vspace{-0.3cm}
  \label{tab:headings}
  \centering
  \resizebox{1.0\textwidth}{!}{
  \begin{tabular}{>{\arraybackslash}m{2cm} | >{\centering\arraybackslash}m{1.5cm}  | >{\centering\arraybackslash}p{1cm} | >{\centering\arraybackslash}p{1cm}}
    \toprule
    \multirow{2}{*}{Method} & \multirow{2}{*}{Backbone} & {Params} & mAP \\
     &  & {(M)} & (\%) \\
    \midrule
    R3Det~\cite{yang2021r3det} & R152  & 76.76 & 76.47 \\
    SASM~\cite{hou2022shape} & RX101  & 58.01 & 79.17 \\
    S$^{2}$ANet~\cite{han2021align} & R50  & 38.83 & 79.42  \\
    ReDet~\cite{han2021redet} & ReR50  & 55.82 & 80.10  \\
    $\text{R3Det}_{\text{\tiny GWD}}$~\cite{yang2021rethinking} & R152  & 76.76 & 80.19 \\
    RTMDet~\cite{lyu2022rtmdet} & R50 & 52.33&80.54 \\
    $\text{R3Det}_{\text{\tiny KLD}}$~\cite{yang2021learning} & R152  & 76.76 & 80.63 \\
    AOPG~\cite{cheng2022anchor} & R50 & 41.95 & 80.66  \\
    KFIoU~\cite{yang2022kfiou} & Swin-T &  60.52& 80.93 \\
    RVSA~\cite{wang2022advancing} & ViTAE-B  & 114.37 & 81.24 \\
    YOLO-v8~\cite{Jocher_Ultralytics_YOLO_2023} & YOLOv8x & 69.5 & 81.36 \\
    \midrule
    \multirow{3}{*}{\parbox{2cm}{Oriented \\ R-CNN~\cite{xie2021oriented}}}
    & R50 & 41.37 &  80.62 \\
    & {$\text{R50}_{\text{ARC}}$} 
     & 75.06 & {81.77}  \\
    & \cellcolor{lightgray!30}{\bf $\text{R50}_{\text{GRA}}$} 
    & \cellcolor{lightgray!30}{\bf 41.65} & \cellcolor{lightgray!30}{\bf 81.93} \\
    \bottomrule
  \end{tabular}
  
  }
% \vskip -0.2in

\label{tab:ms}
\end{minipage}\hfill
\begin{minipage}{0.47\textwidth}
\centering

\caption{{\bf Experiment results on the DOTA-v2.0 dataset.} Our method illustrates its strong performance in detecting tiny objects, achieving SOTA results on the DOTA-v2.0 dataset. $\dagger$ denotes training for 40 epochs.}
% \vspace{-0.3cm}
  \label{tab:headings}
  \centering
  \resizebox{1.0\textwidth}{!}{
  \begin{tabular}{>{\arraybackslash}m{2cm} |>{\centering\arraybackslash}m{1.7cm} |>{\centering\arraybackslash}m{1.75cm}  }
    \toprule
    Method & Backbone &  mAP(\%) \\
    \midrule
    SASM~\cite{hou2022shape} & R50 & 44.53 \\
    R3Det~\cite{yang2021r3det} & R50 &  47.26 \\
    FR-OBB~\cite{ren2015faster} & R50 & 47.32 \\
    FCOS-O~\cite{detector2022fcos} & R50 &  48.51  \\
    Ori-Rep~\cite{li2022oriented} & R50 & 48.95  \\
    ATSS-O~\cite{zhang2020bridging} & R50 &  49.57  \\

    S$^{2}$ANet~\cite{han2021align} & R50 &  49.86  \\
    
    \midrule
    \multirow{3}{*}{DCFL~\cite{xu2023dynamic}}
    & R50   &  51.57  \\
    & R50$\dagger$   &  55.08  \\
    & ReR101$\dagger$   &  57.66 \\
    
    \midrule
    \multirow{4}{*}{\parbox{2cm}{Oriented \\ R-CNN~\cite{xie2021oriented}}}
    & R50 &   53.28 \\
    & {$\text{R50}_{\text{ARC}}$} 
    &  {55.91}  \\
    & \cellcolor{lightgray!30}{\bf $\text{R50}_{\text{GRA}}$} 
    & \cellcolor{lightgray!30}{\bf 56.63} \\
    & \cellcolor{lightgray!30}{\bf $\text{R50}_{\text{GRA}}$ $\dagger$} 
    & \cellcolor{lightgray!30}{\bf 57.95} \\
    \bottomrule
  \end{tabular}
  }
% \vskip -0.2in

\label{tab:dota20}
\end{minipage}

\end{table}

\subsection{Main Results}

\textbf{Effectiveness on various detectors}. We conduct experiments on a variety of popular detectors, including both single-stage methods (Rotated RetinaNet~\cite{lin2017focal}, R3Det~\cite{yang2021r3det}, S$^{2}$ANet~\cite{han2021align}) and two-stage approaches (Rotated  FasterR-CNN~\cite{ren2015faster}, CFA~\cite{guo2021beyond}, Oriented R-CNN~\cite{xie2021oriented}). The results on the DOTA-v1.0 dataset are shown in \cref{tab:main}. In the backbone column, R50 stands for ResNet-50~\cite{he2016deep}, $\text{R50}_{\text{ARC}}$ is the backbone network that replaces the $3\times3$ convolution in the last three stages of ResNet-50 with the ARC~\cite{pu2023adaptive} module and $\text{R50}_{\text{GRA}}$ is the backbone network that replaces the $3\times3$ convolution in the last three stages of ResNet-50 with GRA module.  Experimental results illustrate the effectiveness and compatibility of the proposed method on various frameworks. At the same time, the number of parameters is significantly reduced compared with ARC~\cite{pu2023adaptive}. 
% Compared with the regular ResNet50 \cite{he2016deep}, for single stage methods, our structure can improve the mAP of Rotated RetinaNet~\cite{lin2017focal} by 3.77\%, R3Det~\cite{yang2021r3det} by 3.23\%, S$^{2}$ANet~\cite{han2021align} by 1.85\%. For two-stage methods, our structures can improve Rotated Faster RCNN~\cite{ren2015faster} by 2.05\%, CFA~\cite{guo2021beyond} by 4.6\%, Oriented R-CNN ~\cite{xie2021oriented} by 1.82\%.
\cref{tab:hrsc} further shows the robustness of our method on the HRSC dataset
, evidencing the mAP improvement of 1.27\% for Rotated RetinaNet~\cite{lin2017focal}, 2.32\% for S$^{2}$ANet~\cite{han2021align}, and 1.90\% for Oriented R-CNN~\cite{xie2021oriented} compared with the regular ResNet50. On HRSC dataset, our model can also outperform various previous SOTA methods. 

\textbf{Performance under multi-scale training and testing strategies}. We also assess the performance of GRA under multi-scale training and testing strategies on the DOTA-v1.0 dataset, which is shown in \cref{tab:ms}. Under this setting, various data augmentation methods are applied, which is more suitable for larger models to avoid under-fitting during training. On the other hand, our lightweight GRA still illustrates highly competitive performance.

\textbf{Performance on DOTA-v2.0 dataset}. Compared with the DOTA-v1.0 dataset, the DOTA-v2.0 dataset contains much more tiny objects, which are more challenging to detect. GRA also illustrates its strong performance as shown in \cref{tab:dota20}. Using Oriented R-CNN as the detector and training for 40 epochs, we achieve the SOTA mAP of 57.95\%, which outperforms all previous methods.  

\textbf{Performance under different pretraining strategy}. 
Training a detector typically involves an extensive pre-training phase for its backbone on the ImageNet dataset, followed by further training on task-specific datasets such as DOTA for oriented object detection \cite{tang2024mind}. Previous methods such as ARC~\cite{pu2023adaptive} and LSKNet \cite{li2023large} are limited to a from-scratch pre-training approach for the backbone, which consumes considerable computational resources and time.
In contrast, our GRA framework simply adds several additional modules onto existing foundational models (such as the standard ResNet), without altering their internal structures. This characteristic allows our method to leverage pre-trained weights of standard models that are publicly available, eliminating the need for resource-intensive training from scratch. During the pre-training phase, we load the weights of a standard pre-trained ResNet model, focusing solely on training the newly integrated GRA modules. Meanwhile, other parts of the backbone (i.e., the standard ResNet components) remain frozen. Experimental results (\cref{tab:pretrain}) demonstrate that directly training a randomly initialized model on the DOTA dataset leads to extremely poor performance (35.51\% mAP). However, by loading the publicly available pre-trained weights of ResNet and training only the GRA modules during the pre-training phase, the final trained detector still achieves satisfactory performance (77.39\% mAP). We believe that the flexibility of GRA, further underscores its applicability to larger models, enhancing its utility and adaptability in the field of oriented object detection.

\setlength\tabcolsep{6pt}
% \vskip -0.2in
\begin{table}[ht]
  \caption{\textbf{Results under different pre-training strategies.} \includegraphics[width=0.8em]{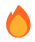} denotes training from scratch. \includegraphics[width=0.8em]{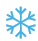} denotes loading the public available pre-trained weight and freezing the parameters during training. The Oriented-RCNN is used as the detector, other experiment settings follow \cref{tab:main}. }
  % \vspace{-0.3cm}
  \label{tab:headings}
  \centering
  \resizebox{0.75\textwidth}{!}{
  \begin{tabular}{l I c |c |c }
    \specialrule{1pt}{1pt}{1pt}
    \multirow{2}{*}{Pretraining Stragegy} & \multirow{2}{*}{No Pretraining  }& ResNet\includegraphics[width=0.8em]{fig/flame.jpg}  & ResNet\includegraphics[width=0.8em]{fig/snow.jpg}    \\
    
    & & GRA module\includegraphics[width=0.8em]{fig/flame.jpg} & GRA module\includegraphics[width=0.8em]{fig/flame.jpg} \\
    
    \midrule
    mAP(\%) & 35.51 & 77.64 & 77.39 \\
    \specialrule{1pt}{1pt}{1pt}
  \end{tabular}
  }
% \vskip -0.2in

\label{tab:pretrain}

\end{table}

\subsection{Ablation Study}
\textbf{Number of groups.} An ablation study was conducted to assess the impact of varying the number of groups within the GRA module (\cref{tab:ablation_number_groups}). Within a certain range, increasing the number of groups results in the angle generator of the Group-wise Rotating module predicting more angles. Consequently, this enables the network to capture finer details of rotating objects with diverse orientations. 
Besides, increasing the number of groups will not lead to a significant rise in parameters and FLOPs. 
% This is attributed to the fact that $n$ can only affect the parameter and FLOPs of two fully connected layers in the Angle Generator and the number of convolution layers in the Group-wise Attention module, constituting only a small portion of the entire network.

% Experimental results indicate that performance improvement correlates with an increase in the number of groups within a certain range. This observation suggests that for objects within an image that exhibit similar orientations, a large number of parameters are not a prerequisite for effective detection. Specifically, when the grouping is set to $n=16$, meaning the parameter of the convolution is divided into 16 distinct groups, each group utilizes merely 1/16 of the total parameters. Despite this significant reduction, the detection outcomes remain notably satisfactory. This phenomenon underscores the criticality of devising strategies that effectively capture angular information in oriented object detection tasks, rather than relying on an increase in parameter count.

\begin{table}[ht]
\centering
\begin{minipage}{0.48\textwidth}
\centering
% \vskip -0.2in
\caption{\textbf{Ablation studies on the number of groups} $n$. The Oriented-RCNN is used as the detector, other experiment settings follow \cref{tab:main}.}
% \vspace{-0.3cm}
\resizebox{\textwidth}{!}{%
\begin{tabular}{c|c|c|c|c}
\toprule
Backbone & $n$ & Params (M) & FLOPs (G) & mAP (\%) \\ 
\midrule
$\text{R50}$ & - & 23.51 & 171.5 & 75.81 \\
$\text{R50}_{\text{GRA}}$ & 2 & 23.56 & 172.4 & 76.82 \\
$\text{R50}_{\text{GRA}}$ & 8 & 23.61 & 172.4 & 77.27 \\
$\text{R50}_{\text{GRA}}$ & 16 & 23.67 & 172.5 & 77.41 \\
$\text{R50}_{\text{GRA}}$ & 32 & 23.79 & 172.7 & \textbf{77.63} \\
\bottomrule
\end{tabular}%
}
% \vskip -0.25in
\label{tab:ablation_number_groups}
\end{minipage}\hfill
\begin{minipage}{0.47\textwidth}
\centering
% \vskip -0.2in
\caption{\textbf{Ablation studies on different parts of GRA module.} The Oriented-RCNN is used as the detector, other experiment settings follow \cref{tab:main}.}
% \vspace{-0.3cm}
\resizebox{\textwidth}{!}{%
\begin{tabular}{c c |c | c}
\toprule
\multicolumn{2}{c|}{\textbf{Group-wise Rotating}} & {\textbf{Group-wise}} & \multirow{2}{*}{mAP (\%)} \\
Rotating & Weighted by $\boldsymbol{\lambda}$ &\textbf{Attention} & \\
\midrule
\xmark & \xmark & \xmark & 75.81 \\
\cmark & \xmark & \xmark & 76.73 \\
\cmark & \cmark & \xmark & 77.25 \\
\cmark & \cmark & \cmark & \textbf{77.63} \\
\bottomrule
\end{tabular}%
}
% \vskip -0.25in
\label{tab:ablation_lambda}
\end{minipage}

\end{table}

Furthermore, it is observed that with a continuous increase in $n$, there is no significant improvement in performance. This phenomenon can be attributed to the situation where, as $n$ becomes excessively large, each group contains only a few kernels. In such cases, if the differences among the predicted angles for these $n$ groups are significant, the limited number of kernels within each group may fail to adequately capture the features corresponding to the rotated angle. Consequently, the network might predict similar angles for various groups to ensure that the features associated with the particular angle maintain satisfactory quality. This behavior is equivalent to merging several groups into one.

\begin{figure}[t]
	\centering
	\includegraphics[width=1.0 \textwidth]{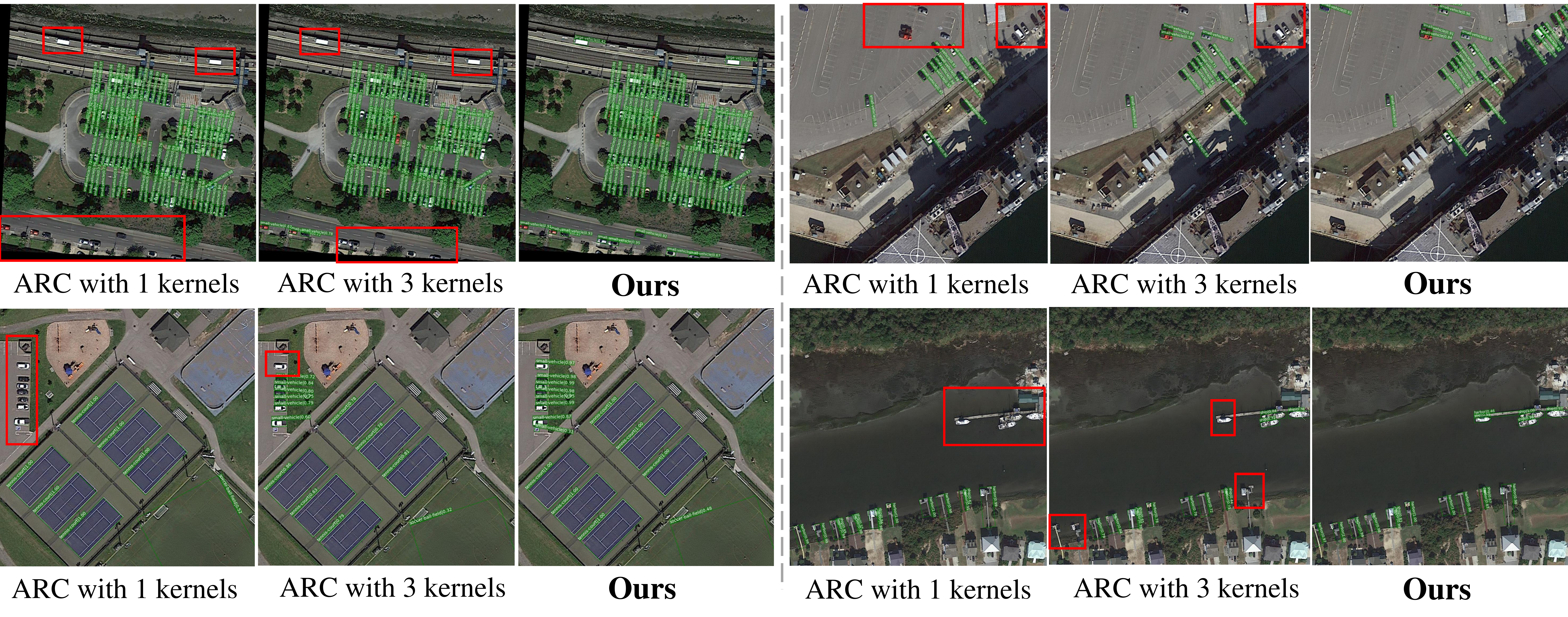} 
	\caption{\textbf{The visualisation of the prediction of ARC~\cite{pu2023adaptive} and our model.}}
	\label{fig:vis}

\end{figure}

\textbf{Each component of GRA.} We also conduct an ablation study to illustrate the effectiveness of different components of GRA (\cref{tab:ablation_lambda}). Group-wise Rotating module and Group-wise Attention module are both of great importance. To be noticed, in the Group-wise Rotating module, each group of kernel is weighted by a learnable scale factor $\lambda_i$, which can measure the importance of different groups. We find that this design can help the model to achieve better performance.

\textbf{Visualization.} 
We provide visualization results to illustrate the effectiveness of our method, which are obtained using the Oriented R-CNN detector on the DOTA-v2.0 dataset under single-scale training. Besides the miss detected object of baseline methods marked in \textcolor{red}{red boxes}, many detected objects detected by ARC with 3 kernels exhibit lower confidence (best-viewed zoom-in).

\textbf{Limitations and future work.} 
Although GRA is simple and outperforms a large number of methods, it is only used to replace the $3\times 3$ convolution kernels in ResNet in our work. Its effectiveness on larger kernels (e.g., $7\times 7$ ) or other convolution-based models (e.g., ConvNeXt) still needs to be explored.

% \begin{figure}[t]
%     \centering
%     \includegraphics[scale=0.27]{fig/vis.pdf}
%     \caption{\textbf{Visualisation of our model's prediction on DOTA v-1.0 dataset}. Our method can learn the more fine-grained features and therefore detect the small objects in the images more precisely. }
%     \label{fig:vis}
%     \vskip -0.2cm
% \end{figure}

\section{Conclusion}

In this paper, we introduce a lightweight yet effective module named Group-wise Rotating and Attention (GRA), which can adaptively capture the fine-grained features of objects with various orientations. Within the Group-wise Rotating module, convolution kernels are segregated into distinct groups, each rotating independently at various angles to dynamically capture the features of oriented objects. The group-wise attention mechanism is introduced to adaptively focus on important regions in the feature, reducing the undesired noises contained in the feature and enhancing the learning of fine-grained features of oriented objects. Comprehensive experimental results across multiple datasets demonstrate that our approach outperforms various state-of-the-art (SOTA) methods. Furthermore, GRA introduces minimal additional parameters, enhancing its versatility and potential for real-world deployment.

\section*{Acknowledgments}
This work was partly supported by Shenzhen Key Laboratory of next generation interactive media innovative technology (No: ZDSYS20210623092001004).
% \input{sec/6_supp}

% ---- Bibliography ----
%
% BibTeX users should specify bibliography style 'splncs04'.
% References will then be sorted and formatted in the correct style.
%
\bibliographystyle{splncs04}
\bibliography{main}
\end{document}